\documentclass[pdflatex,sn-mathphys-num, iicol]{sn-jnl}% Math and Physical Sciences Numbered Reference Style
\usepackage{graphicx}%
\usepackage{multirow}%
\usepackage{amsmath,amssymb,amsfonts}%
\usepackage{pifont}    
\usepackage{amsthm}%
\usepackage{mathrsfs}%
\usepackage[title]{appendix}%
\usepackage{xcolor}%
\usepackage{textcomp}%
\usepackage{manyfoot}%
\usepackage{booktabs}%
\usepackage{algorithm}%
\usepackage{algorithmicx}%
\usepackage{algpseudocode}%
\usepackage{listings}%
\usepackage[table]{xcolor}
\usepackage[T1]{fontenc}
\usepackage{lmodern}
\usepackage{amssymb}          
\usepackage{colortbl}         
\usepackage{soul}             
\usepackage{array}            
\usepackage{verbatim}
\usepackage{tabularx}
\theoremstyle{thmstyleone}%
\theoremstyle{thmstyletwo}%

\theoremstyle{thmstylethree}%
\newcommand{\methodname}{UNBOX}
\newcommand{\xmark}{\ding{55}}
\definecolor{table-gray}{gray}{0.92}

\begin{document}

\title[Article Title]{UNBOX: Unveiling Black-box visual models with Natural-language}


\author*[1]{\fnm{Simone} \sur{Carnemolla}}\email{simone.carnemolla@phd.unict.it}

\author[1]{\fnm{Chiara} \sur{Russo}}\email{chiara.russo1@phd.unict.it}

\author[1]{\fnm{Simone} \sur{Palazzo}}\email{simone.palazzo@unict.it}

\author[2,3]{\fnm{Quentin} \sur{Bouniot}}\email{quentin.bouniot@tum.de}

\author[1]{\fnm{Daniela} \sur{Giordano}}\email{daniela.giordano@unict.it}

\author[2,3]{\fnm{Zeynep} \sur{Akata}}\email{zeynep.akata@tum.de}

\author[1]{\fnm{Matteo} \sur{Pennisi}}\email{matteo.pennisi@unict.it}
\equalcont{Equal supervision.}

\author[1]{\fnm{Concetto} \sur{Spampinato}}\email{concetto.spampinato@unict.it}

\equalcont{Equal supervision.}

\affil[1]{\orgname{University of Catania}}

\affil[2]{\orgname{Technical University of Munich}}

\affil[3]{\orgname{Helmholtz Munich}}

\abstract{
Ensuring trustworthiness in open-world visual recognition requires models that are interpretable, fair, and robust to distribution shifts. Yet modern vision systems are increasingly deployed as proprietary black-box APIs, exposing only output probabilities and hiding architecture, parameters, gradients, and training data. This opacity prevents meaningful auditing, bias detection, and failure analysis. Existing explanation methods assume white- or gray-box access or knowledge of the training distribution, making them unusable in these real-world settings.

We introduce \textbf{\methodname{}}, a framework for \textbf{class-wise model dissection} under fully data-free, gradient-free, and backpropagation-free constraints. \methodname{} leverages Large Language Models and text-to-image diffusion models to recast activation maximization as a purely semantic search driven by output probabilities. The method produces human-interpretable text descriptors that maximally activate each class, revealing the concepts a model has implicitly learned, the training distribution it reflects, and potential sources of bias.

We evaluate \methodname{} on ImageNet-1K, Waterbirds, and CelebA through semantic fidelity tests, visual-feature correlation analyses and slice-discovery auditing.
Despite operating under the strictest black-box constraints, \methodname{} performs competitively with state-of-the-art white-box interpretability methods. This demonstrates that meaningful insight into a model’s internal reasoning can be recovered without any internal access, enabling more trustworthy and accountable visual recognition systems.}




\maketitle
\section{Introduction}
\label{introduction}
\textit{Can the reasoning of a visual classifier be uncovered when the model is treated as a complete black box, with no access to its architecture, class definitions, training data, or learned weights?} As visual recognition increasingly relies on large proprietary models offered exclusively as inference APIs, this question has become central to trustworthy AI. In such deployments, users only observe output probabilities while all internal computations remain hidden, making it extremely difficult to understand why a classifier behaves the way it does, to diagnose failures, or to assess robustness and fairness in open-world settings.

A substantial body of work has explored the dissection of visual recognition models, aiming to uncover the internal concepts and mechanisms that drive their predictions. Local explanation methods, particularly Natural Language Explanations (NLEs) \cite{park2018multimodal, sammani2023uni, sammani2022nlx, hendricks2018grounding, zellers2018recognition}, focus on per-sample reasoning and often rely on multimodal supervision or VQA-style pipelines. In contrast, global explanation and model-dissection approaches aim to characterize a model’s overall conceptual space or the semantic attributes encoded in its representations. Recent methods such as DiffExplainer \cite{PENNISI2025104559}, DEXTER \cite{carnemolla2025dexter}, and CLIPDissect \cite{oikarinen2023clipdissect} use LLMs, diffusion models, or vision-language embeddings to optimize class-level descriptors, expose spurious features, or assign semantic labels to internal units and neurons. While these approaches provide valuable insight, they rely on access to model internals (weights, gradients, activations) or probing datasets. Other explanation strategies assume knowledge of the training distribution or annotated data. Consequently, none of these techniques can be applied in a black box scenario, where a model is only available through an API with no visibility into its architecture, parameters, or training data.

This gap emphasizes the need for a method that can perform global, class-wise model dissection using only a model’s output probabilities. Meeting this requirement demands a conceptual shift: rather than optimizing in feature space through gradients, which is impossible in black-box settings, we propose explanation as a semantic search problem driven entirely by natural language. Recent advances in textual optimization, such as TextGrad \cite{yuksekgonul2024textgrad}, show that LLM-mediated feedback can approximate gradient signals in non-differentiable systems, suggesting a promising direction for black-box interpretability. Motivated by this insight, we reformulate activation maximization as a gradient-free optimization loop carried out purely in text space. 
Building on this idea, we introduce \methodname{}, a class-wise explanation framework that is entirely \emph{data-free, gradient-free, and backpropagation-free}. \methodname{} discovers human-interpretable textual descriptors that strongly activate a target class using only the classifier’s output probabilities. The method combines a text-to-image diffusion model with an agentic, LLM-driven optimization mechanism that iteratively refines a natural-language descriptor for the target output neuron. A semantic guidance signal directs two cooperating LLM agents: one produces structured feedback and the other updates the descriptor. To ensure stability and prevent collapse onto narrow or transient descriptors, \methodname{} maintains a lightweight global and local optimization context that accumulates consistently high-reward prompts and salient lexical units while also tracking recent refinement steps. Through this combination, activation maximization is reformulated as a semantic optimization problem, enabling global model dissection without access to model architecture, parameters, or training data.
    
We evaluate \methodname{} on ImageNet-1K \cite{deng2009imagenet}, Waterbirds \cite{sagawadistributionally}, and CelebA \cite{liu2015deeplearningfaceattributes} through a comprehensive set of complementary tasks designed to assess fidelity, grounding, and practical utility. To measure semantic faithfulness, we compare our recovered descriptors against ground-truth class names and against descriptors extracted from training images via automated captioning.
To assess practical auditing capability, we use the descriptors to perform slice-discovery analysis on Waterbirds and CelebA, revealing spurious correlations and enabling debiasing despite having no access to training data or model internals. Across all tasks, \methodname{} achieves performance competitive with, and in several cases approaching, methods that rely on weights, gradients, or full training datasets, demonstrating its effectiveness as a fully black-box model-dissection tool.

\vspace{4pt}
To summarize, our main contributions are:
\begin{itemize}
\item A fully data-free and gradient-free framework for global model dissection. \methodname{} reconstructs class-level semantic concepts using only output probabilities, reformulating activation maximization as a text-space semantic optimization problem guided by diffusion-based evidence and agentic LLM reasoning.
\item A principled textual optimization mechanism that enables iterative refinement of descriptors through structured natural-language feedback without access to model weights, gradients, or training data.
\item Extensive evaluation across semantic, visual, and fairness-oriented tasks, showing that \methodname{} produces descriptors that are coherent, visually grounded, and effective for downstream auditing, including slice discovery and debiasing, despite operating in the strictest black-box setting.
\item Human validation of interpretability, demonstrating that the recovered descriptors are accurate, meaningful, and competitive with those produced by leading white-box explainability methods.
\end{itemize}

\section{Related work}
A large body of work seeks to interpret visual classifiers by uncovering the internal concepts and mechanisms driving their predictions. Approaches differ along two axes: \textit{intrinsic} methods, which build interpretable models from the outset \cite{elton2020self, yang2023language}, and \textit{post-hoc} methods, which dissect pre-trained models to explain how they arrive at their decisions \cite{selvaraju2017grad, carnemolla2025dexter, oikarinen2023clipdissect, ahn2024unified}.
Both types of methods may provide \textit{local} explanations for individual samples or \textit{global} explanations describing the conceptual structure of the entire model. \methodname{} belongs to the latter category: a global post-hoc method for model dissection and explainability. In this section, we review prior work based on the assumptions they make about model and data access. We first review attribution-based approaches, which offer local insights but no global semantic structure. We then cover concept-based methods, that provide richer explanations but require internal weights, activations, or annotated concepts. Next, we examine global language-driven explanatory methods that perform textual optimization without explicit annotated probes. Finally, we discuss slice-discovery and debiasing frameworks that rely on access to the underlying data.

\noindent \textbf{Attribution-based local explainability.}
Early explainability work focused on attribution methods that highlight which regions of an input image influence a model’s prediction the most. These include feature visualization and saliency-map approaches \cite{simonyan2013deep, selvaraju2017grad, sundararajan2017axiomatic}, improved later by diffusion-based activation maximization techniques such as DiffExplainer \cite{PENNISI2025104559}. Other refinements expanded feature attribution to gradient-based methods \cite{selvaraju2017grad, zeiler2014visualizing, srinivas2019full} and perturbation-based methods \cite{fong2017interpretable, wagner2019interpretable}.

While these methods provide fine-grained insights, they are inherently \textit{local}, often noisy, and require human interpretation to infer the underlying semantic concepts \cite{kim2024discovering}. Model-agnostic surrogates such as LIME \cite{ribeiro2016should} and SHAP \cite{lundberg2017unified} attempt to explain predictions via simpler proxy models but inevitably suffer from information loss.
These methods cannot perform global model dissection and cannot operate in black-box settings, as they require model queries beyond class probabilities, gradients, or feature maps.

\noindent \textbf{Concept-based methods} aim to provide global explanations through human-interpretable semantic "concepts" (e.g., textures, colors, shapes, parts). Traditional approaches such as T-CAV \cite{kim2017interpretability}, T-CAR \cite{crabbe2022concept}, IBD \cite{zhou2018interpretable}, and CRAFT \cite{fel2023craft} require annotated examples or segmentation supervision to identify the relevance of concepts within internal layers.

More advanced variants perform neuron- or filter-level concept assignment. NetworkDissection \cite{bau2017network} and Net2Vec \cite{fong2018net2vec} link feature maps to interpretable units; CAE \cite{gurkan2025concept}, HINT \cite{wang2022hint}, and MILAN \cite{hernandez2021natural} improve concept localization and reduce annotation dependency. Other methods exploit CLIP embeddings: CLIPDissect \cite{oikarinen2023clipdissect} labels neurons with open-ended concepts, CounTEX \cite{kim2023grounding} generates counterfactual explanations, and WWW \cite{ahn2024unified} combines Shapley scores and neuron activation maps to answer the what/where/why of a decision.
Parallel to these, Natural Language Explanations (NLEs) \cite{park2018multimodal, hendricks2018grounding, sammani2022nlx, sammani2023uni, zellers2018recognition} generate textual rationales for predictions but are predominantly local, supervised, and task-dependent.

All these methods require access to model internals (weights, activations, layers, features) or annotated concepts. None can operate with only output probabilities, nor can they fully reconstruct class-level conceptual structures of a black-box model.\\

\noindent \textbf{Global language-driven model dissection}. Several recent methods have explored using textual optimization or generative models for concept-based explanations without explicit annotated probes. TEXplain \cite{asgari2024texplain}, GIFT \cite{zablocki2024gift}, and DEXTER \cite{carnemolla2025dexter} generate class-level descriptors through textual or multimodal optimization. However, these methods still depend on gradients, internal representations, or feature activations, and thus remain incompatible with strict black-box settings.

\methodname{} advances beyond these approaches by performing global, class-wise textual optimization without any access to data, weights, gradients, or internal activations.\\

\noindent \textbf{Slice discovery and debiasing.} Slice discovery seeks to identify coherent subsets of data where a model fails systematically, often as a result of spurious correlations or hidden stratification \cite{ghosh2025ladder}. Modern approaches such as B2T \cite{kim2024discovering} and LADDER \cite{ghosh2025ladder} rely on vision–language models or retrieved text to infer slices that lack explicit annotations. Distributionally Robust Optimization (DRO) \cite{sagawadistributionally} and its refinements \cite{nam2020learning, sohoni2020no, liu2021just, zhang2022correct} are then used to mitigate worst-group errors.

Although slice discovery and debiasing are not the primary objectives of \methodname{}, we show that the descriptors our method uncovers can be effectively used in slice-sensitive analysis and debiasing, performing competitively with approaches that require data access or model internals.

\begin{figure*}[h]
    \centering
    \includegraphics[width=0.8\textwidth]{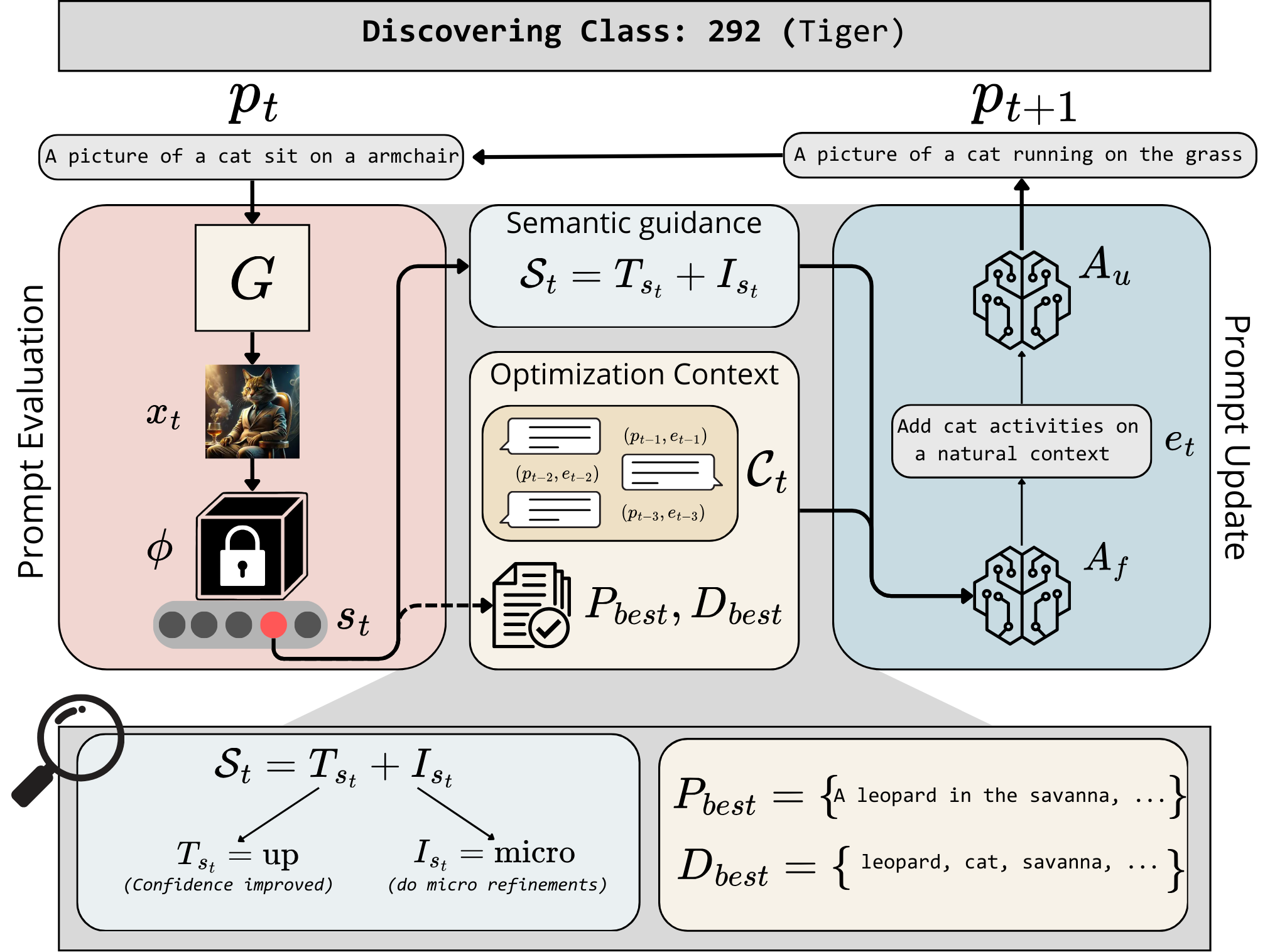}
    \caption{\textbf{Overview of the \methodname{} framework.} Given a target class j and a black-box classifier $\phi$, \methodname{} reformulates activation maximization as a semantic optimization process carried out entirely in text space. The method iterates through three main components. \textbf{(1) Prompt Evaluation (left):} a natural-language prompt $p_t$ is rendered into an image $x_t = G(p_t)$ by a text-to-image diffusion model, which is then evaluated by the classifier to obtain the target-class score $s_t = \phi(x_t)_j$. \textbf{(2) Optimization Context (center):} the scalar feedback $s_t$ is transformed into a semantic guidance signal through trend $T_t$ and intensity $I_t$, which together define the textual loss $\mathcal{S}_t$. This module also maintains a Local Optimization Context $C_t$ capturing recent prompt–feedback pairs, and a Global Optimization Context $(P_{\text{best}}, D_{\text{best}})$ that accumulates high-scoring prompts and persistent lexical concepts. \textbf{(3) Textual Optimization (right):} two cooperating LLM agents perform prompt refinement: the Feedback Agent $A_f$ interprets the semantic signal and optimization context to generate structured feedback $e_t$, which the Updater Agent $A_u$ applies to produce the next prompt $p_{t+1}$. The loop repeats until convergence, yielding a ranked set of human-interpretable descriptors that characterize the target class under strict black-box constraints.}
    \label{fig:method}
\end{figure*}

\section{The \methodname{} Method}
\label{method_intro}
\methodname{} performs class-wise model dissection by searching directly in
\emph{text space} for a natural-language prompt that maximally activates a
target output neuron of a black-box classifier. The method is organized into
two complementary components. First, the \emph{semantic optimization mechanism}
interprets the temporal evolution of the classifier’s output probabilities and
translates this signal into structured natural-language instructions that
iteratively refine the prompt. Second, the \emph{global and local optimization
context} stabilizes this refinement process by accumulating high-reward
semantic cues and by preserving short-range optimization history, thereby
providing both global coverage and local consistency. Together, these
components enable \methodname{} to recover a coherent and comprehensive textual
description of the concept encoded by the target class.
The overall pipeline of \methodname{} is shown in Alg.\ref{alg:unbox} and Fig.\ref{fig:method}.

\subsection{Semantic Optimization Mechanism}
\label{sec:method}

\begin{algorithm}[t]
\caption{Semantic optimization procedure used by \methodname{} to recover class-level descriptors from a black-box classifier.}
\label{alg:unbox}
\begin{algorithmic}[1]

\Require Classifier $\phi$, generator $G$, target class $j$
\Require Agents $A_f$ (Feedback), $A_u$ (Updater)
\Require Global Context $P_{\text{best}}, D_{\text{best}} \gets \emptyset$
\State Initialize prompt $p_0$

\For{$t = 0$ to $T$}
    \State Generate image $x_t \gets G(p_t)$
    \State Score $s_t \gets \phi(x_t)_j$
    \State Compute trend $T_t$ and intensity $I_t$
    \State Build semantic signal $S_t$
    \If{$s_t \ge \tau_{\text{best}}$}
        \State Add $(p_t, s_t)$ to $P_{\text{best}}$
        \State Extract lexical units and update $D_{\text{best}}$
    \EndIf
    \State $e_t \gets A_f(p_t, S_t, P_{\text{best}}, D_{\text{best}})$
    \State $p_{t+1} \gets A_u(p_t, e_t)$

\EndFor

\State Sort $D_{\text{best}}$ by activation score
\State \Return top-$k$ descriptors from $D_{\text{best}}$
\end{algorithmic}
\end{algorithm}

\methodname{} performs class-wise model dissection by iteratively refining a textual description so that, when rendered by a generative model, it maximally activates a chosen output neuron on a target classifier. The entire optimization takes place in \emph{semantic space}, i.e., the space of natural-language descriptors, because the classifier exposes neither gradients, architecture, nor training data. Formally, let 
\[
\phi : \mathcal{X} \to [0,1]^C
\]
be the classifier treated as a black box, and let $j \in \{1,\dots,C\}$ denote the target class. At iteration $t$, the method maintains a textual descriptor $p_t$, which is converted into an image through a pretrained text-to-image diffusion model $G$:
\[
x_t = G(p_t).
\]
The synthetic image is then evaluated by the classifier, which returns the
activation of the target output unit:
\[
s_t = \phi(x_t)_j.
\]
This scalar represents the model’s confidence that the generated sample
belongs to class $j$, and is the \emph{only} numerical feedback
available. 
The challenge is therefore to transform this single
probability value into an
optimization signal that can guide how the textual descriptor $p_t$ should be
updated. \methodname{} addresses this by interpreting the temporal evolution
of $s_t$ through two continuous quantities, \emph{trend} and
\emph{intensity}, which summarize the direction and magnitude of
progress, respectively.

\emph{Trend} captures whether the descriptor is improving with respect to the target class. It compares the current score $s_t$ to an exponential moving average of past scores,
\[
\bar{s}_{t-1} = \alpha s_{t-1} + (1-\alpha)\bar{s}_{t-2},
\]
and defines the deviation
\[
T_t = s_t - \bar{s}_{t-1}.
\]
Positive values indicate upward progress, negative values indicate deterioration, and near-zero values indicate stagnation. We discretize this continuous quantity into a semantic category:
\[
\text{trend\_state}(T_t) =
\begin{cases}
\texttt{up}, & T_t > \varepsilon,\\[4pt]
\texttt{down}, & T_t < -\varepsilon,\\[4pt]
\texttt{flat}, & |T_t| \le \varepsilon,
\end{cases}
\]
with a tolerance $\varepsilon > 0$. 
This categorical state acts as a symbolic cue for the language model, indicating the direction in which the descriptor should evolve.

\emph{Intensity} measures how far the descriptor is from producing a high activation and determines how strong the next update should be. It is defined as a margin-based proximity function:
\[
I_t = \max\{0,\ \tau_{\mathrm{high}} - s_t\},
\]
with $\tau_{\mathrm{high}}$ being a hyperparameter.
A large $I_t$ indicates the prompt is far from saturating the target unit, while small values indicate closeness to an optimal descriptor. As with trend, intensity is discretized when communicating with the language model:
\[
\text{intensity\_state}(I_t) \in \{\texttt{strong},\ \texttt{moderate},\ \texttt{micro}\},
\]
corresponding respectively to coarse, mid-scale, or fine-grained modifications to the current descriptor. These categories loosely resemble the effect of the learning rate in gradient-based methods, although all updates occur purely in linguistic form.

The discretized trend and intensity values
are used to 
index
two lookup tables containing short natural-language templates,

mapping each semantic state to a
corresponding textual instruction.

Selecting
the appropriate entries from these tables and concatenating them yields the
semantic guidance signal \(\mathcal{S}_t\) used by the LLM agents:
\[
\mathcal{S}_t
=
\mathbf{A}\big[\text{trend\_state}(T_t)\big]
\;\Vert\;
\mathbf{B}\big[\text{intensity\_state}(I_t)\big],
\]
where \(\mathbf{A}\) and \(\mathbf{B}\) denote the trend and intensity lookup
tables, respectively, and \(\Vert\) denotes string concatenation.

Once the semantic signal $\mathcal{S}_t$ is composed, it drives the update of the
descriptor through two cooperating LLM agents that operate in sequence. The
first is the \emph{Feedback Agent} $A_f$, which receives the current prompt
$p_t$ together with $\mathcal{S}_t$ and produces a structured natural-language
critique $e_t$ indicating how the descriptor should be modified to better
reflect the concept associated with the target output neuron $j$. The second is
the \emph{Updater Agent} $A_u$, which integrates this critique and returns a
refined prompt:
\[
p_{t+1} = A_u\big(p_t,\ e_t = A_f(p_t,\mathcal{S}_t)\big).
\]

\begin{figure*}[h]
    \centering
    \includegraphics[width=1\textwidth]{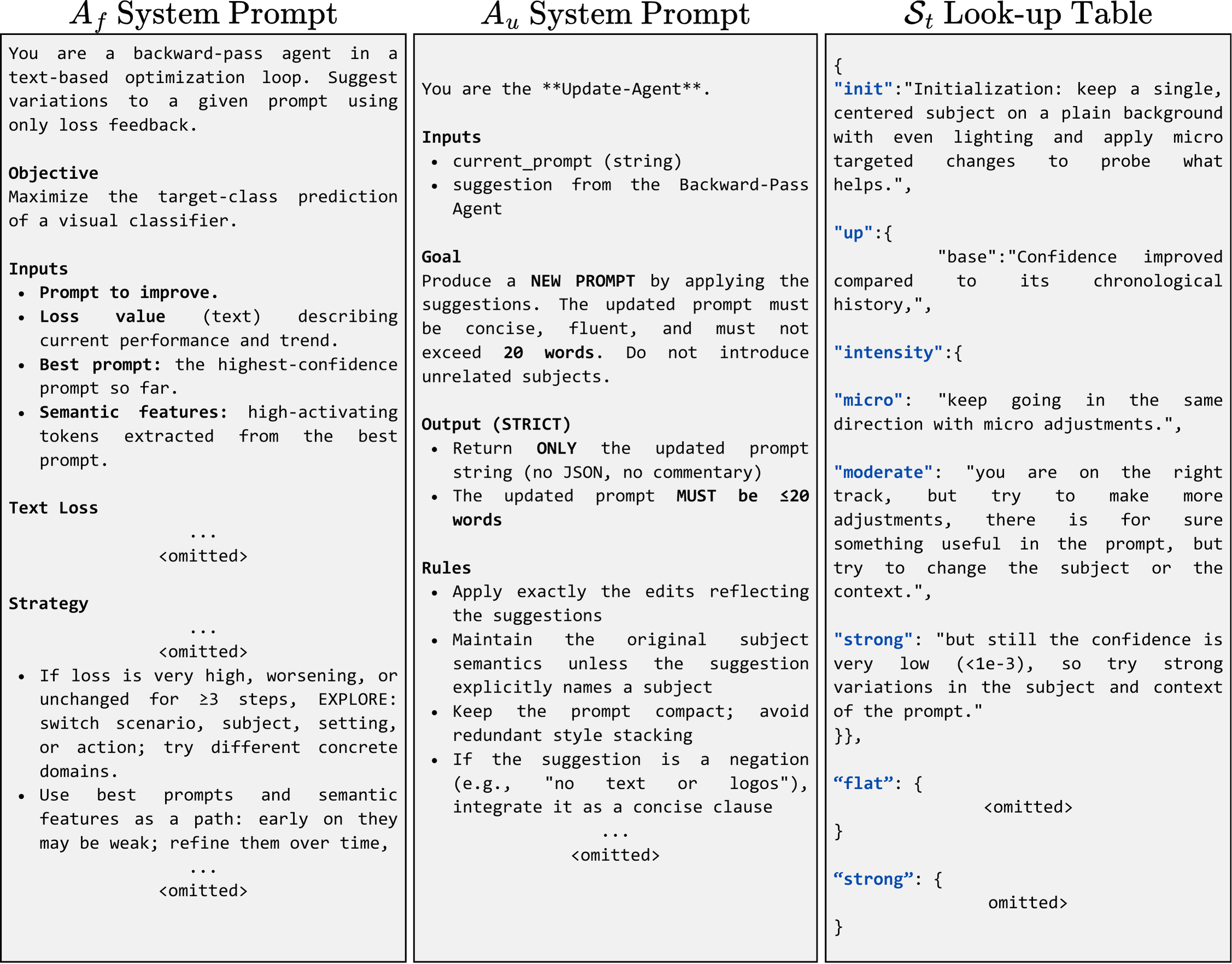}
    \caption{Excerpts of the system prompts for $A_f$ and $A_u$, along with the look-up table used to retrieve the text associated with $T_{s_t}$ and $I_{s_t}$ for the formulation of $\mathcal{S}_t$.}
    \label{fig:system_prompts}
\end{figure*}

Figure~\ref{fig:system_prompts} illustrates the system prompts that govern $A_f$
and $A_u$, together with the lookup table that converts the trend and intensity
states into the textual components of $\mathcal{S}_t$. The prompt of $A_f$
specifies its objective (maximizing the activation of class $j$), the inputs it
receives (current descriptor, semantic signal, and contextual cues from
high-scoring lexical units), and the strategy for producing concise and
actionable edits. The prompt of $A_u$ defines how these edits must be applied,
enforcing constraints on linguistic fluency, subject preservation, and prompt
length. The lookup table provides short, standardized fragments encoding the
direction and magnitude of semantic change, ensuring that $\mathcal{S}_t$ is
constructed through a controlled and reproducible mapping rather than an
unconstrained textual heuristic.

Depending on the direction and magnitude encoded in $\mathcal{S}_t$, the update
step may introduce new semantic attributes, suppress misleading or
non-discriminative elements, adjust the global scene configuration, or apply
fine-grained refinements to the existing description. In this way,
$\mathcal{S}_t$ modulates the scale and nature of the edit proposed by $A_f$ and
executed by $A_u$, ensuring that each iteration reflects a principled response
to the classifier’s feedback.

\subsection{Global and Local Optimization Context}
The optimization process described above operates entirely in semantic space and is therefore highly non-convex: many intermediate prompts may yield transient spikes in class activation without representing stable or meaningful aspects of the underlying concept. We noticed that relying solely on the semantic signal $\mathcal{S}_t$ leads to brittle behaviour: for instance, promising semantic attributes may be discarded and useful refinements overwritten. Most importantly, the absence of an explicit mechanism to leverage optimization history may cause the process to oscillate or stagnate.

To counteract these effects, \methodname{} maintains two complementary structures that summarize information across distinct temporal scales: (i) a \textit{Global Optimization Context}, which aggregates high-scoring prompts and their most relevant lexical units, and (ii) a \textit{Local Optimization Context}, which stores the most recent prompt--feedback interactions. These structures provide high-level semantic cues to the LLM agents without modifying the mathematical optimization signals (trend and intensity).\\ 

\noindent\textbf{Global Optimization Context.}
This component stores all prompts that achieve an activation above a high-confidence threshold $\tau_{\text{best}}$. Whenever $s_t \ge \tau_{\text{best}}$, the corresponding prompt $p_t$ is added to a set $P_{\text{best}}$. After optimization, this set is ranked by activation score, yielding a collection of the most effective prompts discovered for the target class.

To uncover finer-grained semantic cues, each selected prompt $p_t$ is decomposed into its lexical units $\{w_i\}$. After discarding items that do not constitute meaningful standalone semantic units (e.g., stop-words, conjunctions, determiners, auxiliary verbs), the remaining lexical items are treated as candidate semantic concepts. Each is independently evaluated by synthesizing an image from its single-word prompt:
\[
x_{w_i} = G(w_i), \qquad s_{w_i} = \phi(x_{w_i})_j.
\]
A concept set $D_{\text{best}}$, of maximum size $m$, stores the $w_i$ terms with the largest classifier activation $s_{w_i}^j$ obtained up to the current iteration, resulting
in a list of atomic semantic cues that strongly correlate with the targeted class.
These cues represent persistent high-reward concepts that should continue to influence future refinements even if they do not appear in the current descriptor.

This mechanism captures the multifaceted nature of class representations. For example, the class \emph{dog sled} may be activated by \textit{huskies}, \textit{snow}, \textit{sled}, or \textit{people in winter clothing}. A purely iterative update may converge to only one such attribute; the Global Optimization Context preserves the full semantic spectrum uncovered during exploration.\\

\noindent\textbf{Local Optimization Context.}
Alongside global structure, \methodname{} maintains a short sliding record of the most recent prompt-feedback interactions. This local context makes the optimization trajectory explicit: it prevents reintroducing recently rejected prompts and highlights refinements that have consistently improved the activation score over the last few steps.

Formally, the context $C_t$ stores a fixed-length sequence of recent pairs
\[
\{(p_{t-k},\, e_{t-k})\},
\]
where each $p_{t-k}$ is a past descriptor and each
\[
e_{t-k} = A_f(p_{t-k},\, \mathcal{S}_{t-k})
\]
is the critique generated by the Feedback Agent. By exposing $C_t$ to $A_f$, the system can amplify promising semantic directions, avoid short-term oscillations, and achieve smoother progression through the high-dimensional semantic search space.

Both the Global Optimization Context $(P_{\text{best}}, D_{\text{best}})$ and the Local Optimization Context $C_t$ are supplied to the LLM agents as auxiliary guidance. This yields a context-aware update rule of the form:
\[
p_{t+1} =
A_u\!\Big(
    p_t,\;
    A_f\big(p_t,\; \mathcal{S}_t,\; C_t,\; P_{\text{best}},\; D_{\text{best}}\big)
\Big).
\]

Iterations continue until the score evolution indicates convergence or a fixed iteration budget is reached. At that point, the set of best lexical units ($D_{best}$) is ranked by activation score, producing a collection of the most effective prompts discovered for the target class.

\section{Experiments}

We evaluate \methodname{} along three complementary dimensions that reflect the core objectives of black-box model dissection. First, we assess \emph{Semantic Fidelity}, determining whether the textual descriptors produced by \methodname{} are sufficiently informative to identify the correct class using only the classifier's output probabilities. Second, we examine \emph{Latent Training Semantics Recovery}, measuring the extent to which the recovered descriptors reveal the semantic structure implicitly encoded in the model’s training distribution and align with its visual decision cues. Third, we evaluate \emph{Bias Discovery}, testing whether \methodname{} can expose spurious correlations and support slice identification on established robustness benchmarks.

Together, these evaluations address the central question of black-box dissection: starting solely from output probabilities, can we infer what the model recognizes, what latent semantics it has internalized, and what systematic biases it relies upon?

\label{experiments_intro}

\subsection{Experimental setup}
Our experiments are designed to answer three questions: 
(1) how accurately \methodname{} can recover the concept associated with a target class using only output probabilities (Semantic Fidelity); 
(2) whether the recovered descriptors reflect the latent semantics encoded in the model's training distribution (Latent Training Semantics Recovery); and 
(3) whether the method can reveal spurious correlations and coherent dataset slices (Bias Discovery). 
All evaluations use established vision benchmarks and compare \methodname{} against the most recent state-of-the-art methods under standardized protocols.\\

\noindent \textbf{Datasets.} For Semantic Fidelity and Latent Training Semantics Recovery, we use the ImageNet-1K validation set, restricted to the 30-class subset introduced by \cite{carnemolla2025dexter}, following the 15 spurious / 15 non-spurious split defined by Salient ImageNet \cite{singla2021salient}.

For Bias Discovery, we use the standard CelebA \cite{liu2015deeplearningfaceattributes} and Waterbirds \cite{sagawadistributionally} benchmarks, as done in recent slice-discovery and debiasing work \cite{yu2025error, kim2024discovering, ghosh2025ladder, nam2020learning, sohoni2020no, liu2021just, zhang2022correct, carnemolla2025dexter}. CelebA contains over 200k face images annotated with 40 binary attributes, while Waterbirds introduces background-related spurious correlations by compositing bird images onto mismatched environments.\\

\noindent \textbf{Models investigated.} We evaluate \methodname{} on two representative and structurally divergent classifiers:
(1) \emph{ResNet50}, a convolutional architecture with strong locality and hierarchical feature extraction, and 
(2) \emph{ViT-B/16}, a transformer-based model that relies on global self-attention with markedly different inductive biases.
These two models were selected not as a limitation of the approach, but to demonstrate that a purely output-driven semantic optimization procedure can generalize across fundamentally different visual processing mechanisms.
Since \methodname{} operates on the output probabilities of a classifier, it can be applied to \emph{any} image classification model, regardless of architecture, modality of internal computation, or access to weights, while maintaining the same level of generality.\\

\noindent \textbf{Competitors.} We compare against CLIPDissect \cite{oikarinen2023clipdissect} and WWW \cite{ahn2024unified}, both of which provide neuron-level explanations but rely on weights and data access, and DEXTER \cite{carnemolla2025dexter}, a grey-box method requiring weights access. This positions \methodname{}---which operates solely from class probabilities---in a strictly more constrained setting. To ensure a fair comparison, we utilize the set of five class lexical units generated by \methodname{} (Sec. \ref{optimization_setup}). For the neuron-centric baselines (CLIPDissect and WWW), we construct an equivalent set by aggregating the top-4 penultimate-layer neuron lexical units and appending the descriptor from the classification head. We strictly adhere to their implementation guidelines, adopting the exact concept vocabularies: WordNet-20K for CLIPDissect and the 80,000 most frequent English words for WWW.  As the probing dataset for both, we employ the same validation set used for the visual grounding evaluation (Tab. \ref{tab:image_zero_shot}).
Unlike competing deterministic methods, DEXTER and \methodname{} are non-deterministic; therefore, we run the optimization three times and report the mean and standard deviation.
For bias and slice-discovery experiments, we compare against the current state-of-the-art methods used in robustness literature, including text-based methods such as B2T \cite{kim2024discovering} and LADDER \cite{ghosh2025ladder}, and with the standard DRO baseline \cite{sagawadistributionally}.\\

\noindent \textbf{Metrics.} Evaluation uses complementary semantic and visual metrics. 
Semantic Fidelity is measured by embedding-based similarity between generated descriptors and ground-truth labels using Sentence Transformers and CLIP. 
Latent Training Semantics Recovery is evaluated through descriptor-to-caption similarity and descriptor-to-image alignment, quantifying how well the recovered semantics reflect the model’s implicit training distribution. 
Bias Discovery is assessed using the standard slice-level accuracy, worst-group accuracy, and spurious-correlation detection metrics used in prior work.

\subsection{Optimization setup}
\label{optimization_setup}
For each class-wise dissection, we allocate a maximum of 1000 optimization steps, 
with early stopping triggered once the visual classifier predicts the target class 
for 10 consecutive generated images. Images are produced using 
\texttt{black-forest-labs/FLUX.1-schnell}~\cite{flux2024} with a single diffusion step, 
while all agents ($A_f$ and $A_u$) operate on \texttt{gpt-oss:120b}~\cite{openai2025gptoss120bgptoss20bmodel}. 
On a system equipped with two H100 GPUs---one handling image synthesis and classifier 
queries, the other dedicated to LLM inference---each optimization step requires 
approximately 15 seconds.

The trend--state tolerance is set to $\varepsilon = 10^{-4}$, and the thresholds 
controlling intensity-state discretization ($\tau_{\text{high}}$) and global-context 
accumulation ($\tau_{\text{best}}$) are set to $10^{-2}$ for standard class-optimization 
experiments. For slice-discovery tasks, which require less sensitivity to small 
probability variations, both thresholds are set to $9 \times 10^{-1}$. The global context 
retains the top $m = 5$ lexical units extracted from high-scoring prompts, while 
the local optimization context stores the previous $k = 10$ prompt--feedback pairs.

Prompt initialization follows the task setting: standard class dissection begins 
from the generic template ``\texttt{a picture of a random object}'', whereas slice 
discovery uses a domain-aware initialization, 
``\texttt{a picture of a [DOMAIN]}'' (e.g., ``\texttt{a picture of a bird}'' for 
Waterbirds).

\subsection{Results}
\subsubsection{Class Semantic Fidelity}
\label{sec:fidelity}
\label{sec:semantic_fidelity}

A fundamental question in model dissection is whether the recovered descriptors truly identify the intended concept associated with a target class. For a black-box setting such as ours, where no gradients, features, or training data
are available, this becomes particularly challenging: the method must infer a class-level semantic description using only the scalar probability returned by the model. An evaluation of \emph{semantic class fidelity} therefore serves as the
most direct test of whether \methodname{} has successfully uncovered the core semantics that the classifier associates with each class.

To assess this property, we compare the descriptors generated by
\methodname{} with the ground-truth ImageNet class labels. An effective dissection method should produce descriptors whose meaning is significantly closer to the correct class label than to unrelated alternatives. Concretely, we concatenate the top--5 descriptors produced for each class into a single
prompt, encode both descriptor and label using T5-XL, and compute their class semantic similarity.

\begin{table}[htbp]
    \centering

    \rowcolors{1}{table-gray}{white}   
    \caption{Class semantic fidelity evaluation: similarity between generated descriptors and ImageNet-1K class labels on three architectures. Columns W and D denote access to model weights and data, respectively.  Higher values indicate stronger alignment with the ground-truth class concept. As DEXTER and \methodname{} are non-deterministic, results are averaged over three runs.}
    \label{tab:labels_zero_shot}

    \renewcommand{\arraystretch}{1.4}
    \begin{tabular}{l c c c c}
        \toprule
        \rowcolor{white} 
        \textbf{Method} & \textbf{W} & \textbf{D} & \textbf{RN50} & \textbf{ViT} \\
        \midrule
        WWW \cite{ahn2024unified}          & \checkmark & \checkmark         & 0.60        & 0.40 \\
        CLIPDissect \cite{oikarinen2023clipdissect}     & \checkmark & \checkmark      & 0.77 & 0.27 \\
        DEXTER \cite{carnemolla2025dexter}         & \checkmark & \xmark             & $0.47\pm0.03$             & $0.34\pm0.01$            \\
        \methodname     & \xmark     & \xmark      & $0.64\pm0.09$ & $0.49\pm0.04$    \\
        Random          & \xmark     & \xmark             & 0.03       & 0.03 \\
        \bottomrule
    \end{tabular}
\end{table}  

We evaluate against three baselines representative of the current state of the
art. WWW and CLIPDissect are white-box methods that require full access to
model weights and a probing dataset; DEXTER is a grey-box method relying on
activation maximization over model weights alone. A random-word baseline is
reported as a lower-bound control.

Results in Tab.~\ref{tab:labels_zero_shot} show that \methodname{} achieves high semantic fidelity across all architectures despite operating under the
strictest constraints. 
On ResNet50, \methodname{} reaches an accuracy of 0.64 exceeding WWW while remaining competitive with CLIPDissect, that, however, requires access to data and models' weights. The advantage becomes even more evident on ViT models: \methodname{} achieves the highest score (0.49), dramatically outperforming CLIPDissect and WWW.

These results indicate that methods relying on internal activations or neuron probes may struggle to generalize across different architectural paradigms, while \methodname{}—driven purely by output probabilities—maintains stable and robust interpretability performance. This demonstrates that semantic
optimization in text space is not only feasible for black-box models but can match or surpass white-box competitors in recovering the core concept encoded by a class.

\subsubsection{Latent Training Semantics Recovery}\label{sec:latent}
\begin{table}[t]
    \centering
    %\small 

    \rowcolors{1}{table-gray}{white}
    
\caption{Alignment between generated descriptors and data-derived semantic attributes.  
Scores quantify the similarity between each method’s descriptors and keywords extracted from captioned training images, reflecting how well the recovered semantics match the distribution the classifier was trained on. Columns W and D denote access to model weights and data, respectively.  Higher values indicate stronger data-grounded semantic recovery. As DEXTER and \methodname{} are non-deterministic, results are averaged over three runs.}    \label{tab:data_der_zero_shot}
    
    \renewcommand{\arraystretch}{1.4}
    
    \begin{tabular}{l c c c c }
        \toprule
        \rowcolor{white} 
        \textbf{Method} & \textbf{W} & \textbf{D}  & \textbf{RN50} & \textbf{ViT} \\
        \midrule
        
        WWW \cite{ahn2024unified}            & \checkmark & \checkmark          & 0.43          & 0.27 \\
        CLIPDissect \cite{oikarinen2023clipdissect}    & \checkmark & \checkmark & 0.73  & 0.30 \\
        DEXTER \cite{carnemolla2025dexter}         & \checkmark & \xmark       & $0.49\pm0.01$  
        & $0.34\pm0.01$     \\
        \methodname     & \xmark     & \xmark             & $0.63\pm0.04$         &  $0.45\pm0.03$   \\
        Random          & \xmark     & \xmark              & 0.03        & 0.03 \\
        
        \bottomrule
    \end{tabular}
\end{table}

\begin{table}[htbp]
    \centering

    \rowcolors{1}{table-gray}{white}
    
\caption{Visual grounding evaluation on ImageNet-1K.  
Generated descriptors are used as textual queries over validation images to assess how strongly they correspond to concrete visual evidence.  Columns W and D denote access to model weights and data, respectively.
Ground-truth labels (GT) serve as an upper bound. As DEXTER and \methodname{} are non-deterministic, results are averaged over three runs. \label{tab:image_zero_shot}}
    
    \renewcommand{\arraystretch}{1.4}
    
    \begin{tabular}{l c c c c  }
        \toprule
        \rowcolor{white} 
        \textbf{Method} & \textbf{W} & \textbf{D} & \textbf{RN50} & \textbf{ViT} \\
        \midrule
        
        GT (Labels)     & -        & -         & 0.90 & 0.90 \\
        GT (Descr.)& \xmark     & \checkmark              & 0.76       & 0.76          \\
        WWW \cite{ahn2024unified}            & \checkmark     & \checkmark         & 0.66       & 0.63          \\
        CLIPDissect \cite{oikarinen2023clipdissect}     & \checkmark     & \checkmark     & 0.85 & 0.36          \\
        DEXTER \cite{carnemolla2025dexter}          & \checkmark     & \xmark         & $0.42\pm0.01$             & $0.36\pm0.02$              \\
        \methodname     & \xmark     & \xmark       & $0.56\pm0.04$ & $0.43\pm0.04$             \\
        Random          & \xmark     & \xmark               & 0.03                 & 0.03          \\
        
        \bottomrule
    \end{tabular}
\end{table}

While semantic fidelity assesses whether \methodname{} can identify the correct class from output probabilities alone, a core objective of model dissection is to determine whether the explanations capture the \emph{semantics the classifier actually learned from its training distribution}. A descriptor may sound plausible or natural-language–coherent without corresponding to attributes genuinely used by the model.
To address this, we evaluate whether the descriptors recovered by \methodname{} reflect (i) concepts present in the training data and (ii) visual patterns that meaningfully influence the classifier’s predictions.

\paragraph{Alignment with data-derived descriptors.}
A single class label is often insufficient to reveal the richness of the underlying data distribution: classes exhibit characteristic co-occurring features, contextual patterns and, in some cases, spurious cues.
To evaluate whether \methodname{} captures this broader semantic structure, we construct a set of data-derived descriptors by captioning 100 training images per class with LLaVA and extracting the most frequent informative keywords (after removing stop-words and non-semantic tokens).
These keywords form a data-grounded description of what appears in the training samples.

For each class, we concatenate the top-5 descriptors produced by \methodname{} and by the data-derived process into two respective prompts, encode them using T5-XL embeddings, and measure cosine similarity.
We compare against DEXTER, CLIPDissect, WWW, and a random-word baseline.

As shown in Tab.~\ref{tab:data_der_zero_shot}, CLIPDissect performs strongly due to its direct access to model weights and data. However, \methodname{} consistently ranks among the top performers despite operating under strict black-box constraints.
On ResNet50 it outperforms WWW; and on ViT—where internal activation probing used by white-box methods becomes less reliable: \methodname{} achieves the highest score overall.
These results indicate that \methodname{} uncovers features that reflect the distribution the classifier was trained on, rather than generating plausible but ungrounded explanations.

\paragraph{Visual grounding on validation images.}
Textual similarity alone cannot guarantee that recovered descriptors correspond to \emph{visually grounded} evidence in the data.
Therefore, we treat each descriptor as a pseudo-label and measure how well it aligns with ImageNet validation images of the corresponding class. This evaluates whether the descriptors encode features that are not only present in the training data but also consistently expressed in the model’s visual decision cues.

An important distinction is that CLIPDissect and WWW both rely on the validation set as probing data for concept extraction, whereas \methodname{} does not access any images during descriptor generation.
Thus, this experiment constitutes a stricter test for our method.

As shown in Tab.~\ref{tab:image_zero_shot}, \methodname{} demonstrates strong grounding performance.
On ResNet50, it nearly closes the gap with WWW; and on ViT, it substantially outperforms CLIPDissect, whose neuron-level probes degrade on transformer architectures.

These findings jointly indicate that \methodname{} recovers descriptors that are both semantically aligned with the model’s training distribution and visually grounded in the data. Despite having no access to model weights, gradients, or training images, the method uncovers latent semantic structure that white-box methods typically access only through internal activations.

\subsubsection{Slice discovery and debiasing}\label{sec:slice}
The final component of our evaluation concerns whether \methodname{} can uncover
\emph{systematic spurious correlations} and identify dataset slices where the classifier fails.
Unlike class-level concept discovery, which targets stable defining features of a class,
slice discovery requires isolating subtle, fine-grained cues—such as backgrounds or demographic attributes—that influence predictions only within specific subsets of samples.

To focus the optimization on relevant variations, we initialize the search with a lightweight
domain prior (e.g., ``a picture of a bird'' for Waterbirds or ``a picture of a celebrity'' for CelebA).
This assumption reflects realistic auditing scenarios where the model’s exact taxonomy may be unknown,
but its application domain is typically evident.

\begin{table*}[htbp]
    \centering
    \small 
    \setlength{\tabcolsep}{4pt}

    \rowcolors{3}{table-gray}{white}
    
    \caption{Comparison of slice discovery and debiasing performance on the Waterbirds and CelebA benchmarks. 
    Metrics include worst-group (\textit{Worst}) and average (\textit{Avg}) accuracy.\label{tab:slice_discovery}}
    
    \renewcommand{\arraystretch}{1.3}
    
    \begin{tabular}{lccccccc}
        \toprule
        \rowcolor{white}
        & & & & \multicolumn{2}{c}{\textbf{Waterbirds}} & \multicolumn{2}{c}{\textbf{CelebA}} \\
        \cmidrule(lr){5-6} \cmidrule(lr){7-8}
        
        \rowcolor{white}
        \textbf{Method} & \textbf{Weights} & \textbf{Data} & \textbf{GT} & \textbf{Worst} & \textbf{Avg} & \textbf{Worst} & \textbf{Avg} \\
        \midrule
        
        ERM       & \checkmark & \checkmark & \xmark      & 62.6$\pm$0.3 & 97.3 & 47.7$\pm$2.1 & 94.9 \\
        LfF \cite{nam2020learning}     & \checkmark & \checkmark & \xmark      & 78.0         & 91.2 & 77.2         & 85.1 \\
        GEORGE \cite{sohoni2020no}  & \checkmark & \checkmark & \xmark      & 76.2$\pm$2.0 & 95.7 & 54.9$\pm$1.9 & 94.6 \\
        JTT \cite{liu2021just}     & \checkmark & \checkmark & \xmark      & 83.8$\pm$1.2 & 89.3 & 81.5$\pm$1.7 & 88.1 \\
        CNC \cite{zhang2022correct}    & \checkmark & \checkmark & \xmark      & 88.5$\pm$0.3 & 90.9 & 88.8$\pm$0.9 & 89.9 \\
        DRO  \cite{sagawadistributionally}    & \checkmark & \checkmark & \checkmark & 89.9$\pm$1.3 & 91.5 & 90.0$\pm$1.5 & 93.3 \\
        LADDER \cite{ghosh2025ladder}  & \xmark      & \checkmark & \xmark      & 92.4$\pm$0.8 & 93.1 & 89.2$\pm$0.4 & 89.8 \\
        DRO B2T \cite{kim2024discovering}  & \xmark      & \checkmark & \xmark     & 90.7$\pm$0.3 & 92.1 & 90.4$\pm$0.9 & 93.2 \\
        DEXTER \cite{carnemolla2025dexter}  & \checkmark & \xmark      & \xmark      & 90.5$\pm$0.1 & 92.0 & 91.3$\pm$0.01 & 91.7 \\
        \methodname & \xmark   & \xmark      & \xmark      & 	88.6$\pm$0.2        & 91.0&    90.01$\pm$0.3         &  90.91   \\
        
        \bottomrule
    \end{tabular}
\end{table*}

Because spurious cues may be averaged out during global optimization,
we additionally extract instance-level descriptors.
Whenever a generated image $x_t$ activates the target class with high confidence,
we pass $x_t$ through a vision–language model (LLaVA) to obtain $k=5$
fine-grained visual descriptors.
Aggregating all descriptors collected throughout the optimization and ranking them by frequency
reveals the dominant slice-defining attributes for each class.
For Waterbirds, for example, this process naturally identifies ``branch'' for landbirds
and ``water'' for waterbirds, corresponding to the well-known background bias.

To validate that these descriptors accurately capture the spurious correlations affecting the classifier,
we incorporate them into a debiasing pipeline following B2T~\cite{kim2024discovering}.
Each descriptor is embedded using CLIP to build textual prototypes, which serve as
pseudo-labelers of the training data.
Images whose visual features align more strongly with the prototype of the opposite class
are annotated as belonging to the conflict slice.
These groups are then used to retrain a classifier with DRO~\cite{sagawadistributionally},
and the resulting worst-group accuracy provides a quantitative measure of the quality of our slice descriptors.

Table~\ref{tab:slice_discovery} reports worst-group accuracy on Waterbirds and CelebA.
On \emph{Waterbirds}, \methodname{} achieves 88.6 worst-group accuracy,
performing competitively with
white-box and data-dependent slice discovery methods, despite using only output probabilities.
On \emph{CelebA}, \methodname{} attains 90.01 worst-group accuracy, matching Oracle DRO and closely approaching DEXTER.

These results show that \methodname{} not only recovers class-defining concepts
and latent training semantics but also exposes the subtle, systematic biases that drive distribution shifts.
Despite operating under strict black-box constraints,
the discovered textual descriptors are sufficiently precise to enable effective debiasing on standard robustness benchmarks.

\subsubsection{Ablation}
\label{ablation}
A key question in understanding the behaviour of \methodname{} is how much of its performance depends on the auxiliary components introduced in Sec. \ref{sec:method}.
While the semantic optimisation mechanism can operate using only trend-based signals, the full method also incorporates two additional elements:
(i) the Global Optimization Context ($P_{best}$, $D_{best}$), which aggregates long-term semantic evidence; and
(ii) the intensity state $I_t$, which modulates the strength of textual refinements.
The ablation study in Table \ref{tab:ablation_study} evaluates the contribution of these components and clarifies whether they are necessary for stable and semantically meaningful convergence.

The evaluation, ran on a subset of 10 classes, uses the same three criteria adopted in the main results:
\begin{enumerate}
    \item descriptor–label similarity ($DL$), assessing Semantic Class Fidelity (Sec. \ref{sec:fidelity});
	\item	descriptor–data similarity ($DD$), assessing alignment with data-derived descriptors as defined in  (Sec. \ref{sec:latent});
	\item	descriptor–image similarity ($DI$), assessing visual grounding on validation images (Sec. \ref{sec:latent}).
\end{enumerate}

When both the Global Optimization Context and the intensity state are disabled, performance is consistently low across all three measures (0.50, 0.30, 0.42).
This shows that trend-based optimisation alone is insufficient and leads to unstable or incoherent descriptors.

Enabling only the intensity state yields partial gains, particularly in descriptor–data similarity (from 0.30 to 0.60).
This indicates that controlling the magnitude of semantic refinements helps reduce oscillations, but does not fully anchor the optimisation to the model’s learned concepts.

When both components are active, performance improves substantially across all metrics (0.90 on descriptor–label and descriptor–data similarity, 0.87 on descriptor–image similarity).
This demonstrates that the Global Optimization Context provides essential long-range semantic structure: it accumulates stable cues, prevents semantic drift, and reinforces refinements that consistently increase the class activation.
Combined with intensity-aware adjustment, these mechanisms enable \methodname{} to consistently recover descriptors that are semantically faithful, grounded in the training distribution, and visually aligned with the classifier’s behaviour under strict black-box constraints.

\begin{table}[h!]
    \centering
    %\small 
    
    \rowcolors{2}{table-gray}{white}
     \renewcommand{\arraystretch}{1.2}
    \setlength{\tabcolsep}{10.5pt}  
    \caption{Ablation study evaluating the contributions of the Global Optimization Context $(P_{best}, D_{best})$ and the intensity state $I_t$ to the overall performance.}
    \label{tab:ablation_study}
       
    \begin{tabular}{ccccc}
        \toprule
        \rowcolor{white} 
        \boldmath$P_{best}, D_{best}$ & \boldmath$I_t$ & \textbf{DL} & \textbf{DD} & \textbf{DI} \\
        \midrule
        
        \xmark & \xmark & 0.50 & 0.30 &  0.42\\ 
        \xmark & \checkmark & 0.50 & 0.60 & 0.40 \\
        \checkmark & \checkmark & 0.90 & 0.90 & 0.87 \\
        \bottomrule
    \end{tabular}
\end{table}

\subsubsection{Emergent Spurious Feature Attribution}\label{sec:emergent_properties}
\begin{figure*}[h]
    \centering
    \includegraphics[width=1\textwidth]{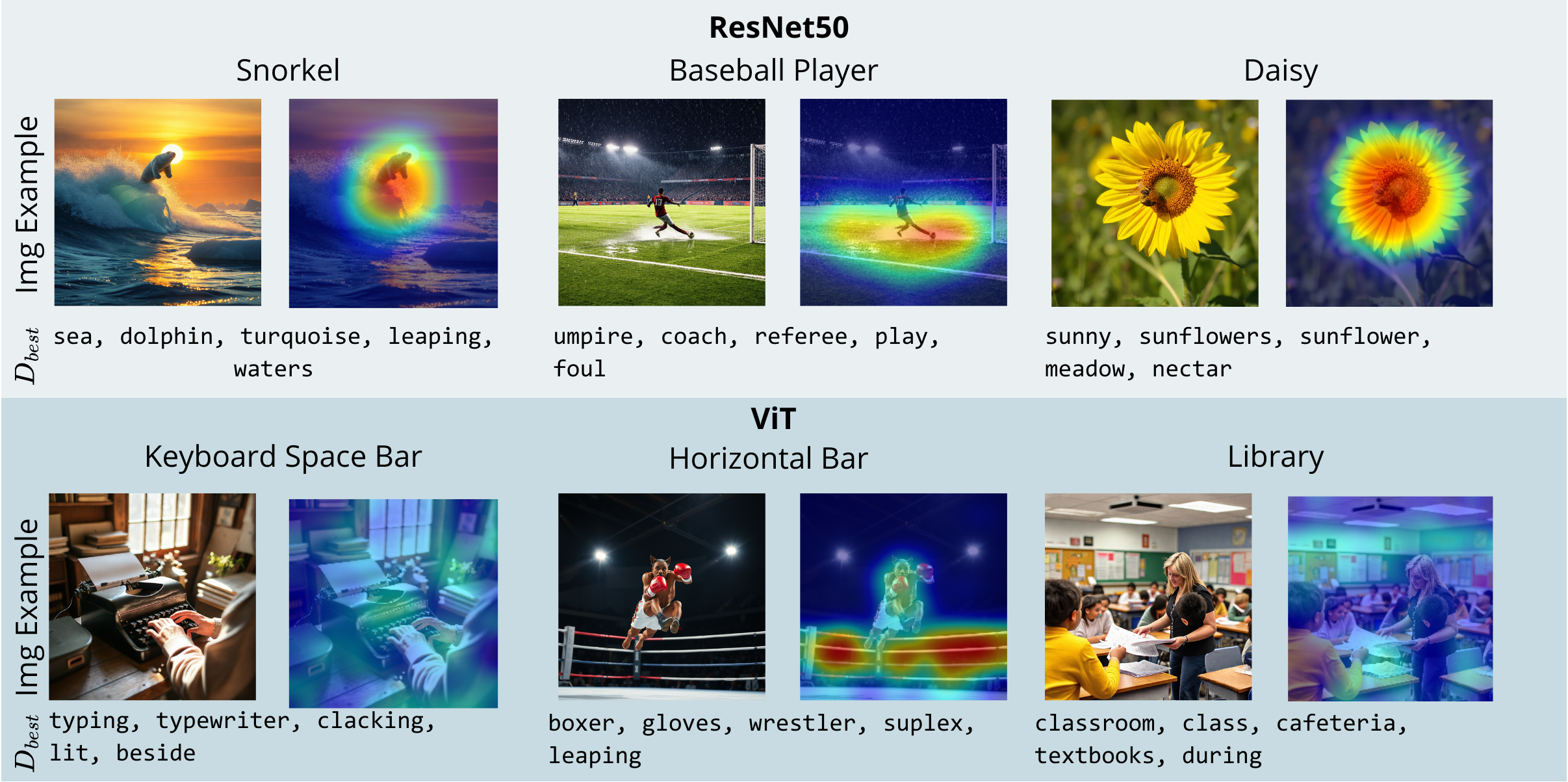}
    \caption{Qualitative examples of spurious feature attribution revealed by \methodname{}.  
For each class, we show generated images that activate the target neuron, the corresponding GradCAM visualizations, and the highest-scoring lexical units in $D_{\text{best}}$. These examples illustrate how non-causal cues (e.g., contextual elements, co-occurring objects, or geometric patterns) can dominate the classifier’s decision process.}
\label{fig:spurious_discovery}
\end{figure*}

A central objective of model dissection is not only to recover the intended semantic concept associated with a class, but also to expose the unintended cues a classifier relies on when making its predictions. Deep models frequently learn to associate target labels with non-causal regularities in the training data, such as backgrounds, co-occurring objects, or simple geometric patterns. Identifying these spurious cues is especially important in a strict black-box setting, where the auditor has no access to training data or internal activations.

\methodname{} naturally reveals such behavior as a byproduct of its optimization process. Because the prompt is refined solely through feedback derived from output probabilities, the procedure gravitates toward whatever visual patterns most strongly activate the classifier. By inspecting the high-confidence generated images, their corresponding GradCAM maps, and the lexical descriptors accumulated in $D_{\text{best}}$, we obtain a direct view of the non-causal features that the model treats as discriminative. Representative examples are shown in Fig.~\ref{fig:spurious_discovery}.

For ResNet50, the \textit{snorkel} class is often triggered by the presence of water, waves, or marine animals rather than by the snorkel apparatus itself. The \textit{baseball player} class displays a similar pattern: environmental cues associated with outdoor sports dominate over features specific to baseball, leading to confusion with classes such as soccer. A related issue arises in fine-grained categories such as \textit{daisy}, where the classifier focuses on background vegetation rather than flower morphology.

For ViT models, the reliance on scene-level or geometric regularities is even more pronounced. The \textit{space bar} class is activated by generic typing scenarios regardless of whether the actual key is visible. The \textit{horizontal bar} class frequently responds to straight-line structures; GradCAM often highlights the edge of an unrelated object whose geometry resembles the target class. Likewise, \textit{library} often activates in indoor educational environments such as classrooms or notice boards rather than canonical library scenes.

These qualitative findings provide important context for the quantitative results in Sections~\ref{sec:semantic_fidelity} and~\ref{sec:latent}. Since the purpose of \methodname{} is to reveal the classifier's actual reasoning process, any dependence on spurious cues should be reflected in the extracted descriptors. When this occurs, the recovered descriptors may diverge from the human-defined class semantics, and such divergence is expected. Lower similarity scores in some tables may therefore indicate genuine model bias rather than a shortcoming of the method. If the classifier strongly associates ``typing at a keyboard'' with the \textit{space bar} class, a descriptor aligned with this association accurately reflects its decision process, even if it deviates from the intended concept.

\section{Limitation and failure modes}
Despite its strong empirical performance, \methodname{} exhibits several limitations inherent to semantic optimization in discrete text space.\\

\noindent \textbf{Sensitivity to low-level or non-semantic visual cues}.
Because the optimization operates over natural-language descriptors, it inherently favors high-level, nameable concepts. When a classifier's decision depends mainly on low-level features—textures, fine-grained color patterns, or local shape primitives—the probability signal $s_t$ often responds only weakly to textual refinements.
A representative failure case is howler monkey: in ImageNet, many images contain dense foliage backgrounds with subtle texture–shape configurations. These cues are difficult to express linguistically, so the optimization receives a weak or noisy trend signal, leading to oscillations and poor convergence. In such settings, the method may fail to recover the true visual evidence used by the classifier.\\

\noindent \textbf{Computational cost of iterative refinement.}
Each optimization step involves two LLM forward passes and one diffusion-model rendering. On a system equipped with two H100 GPUs, a single step takes approximately 15 seconds, implying that full dissection of a class requires non-trivial compute. Although runtime scales with available hardware, this cost remains higher than that of single-pass attribution or gradient-based probing.

These limitations arise from the discrete linguistic search space and the reliance on large generative models for probing. Nonetheless, for the majority of classes, \methodname{} successfully uncovers stable, human-interpretable descriptors under strict black-box constraints.

\section{Conclusion}
\label{conclusion}

We introduced \methodname{}, a framework for class-wise model dissection that operates under the strictest black-box constraints: access is limited solely to output probabilities, with no gradients, features, model weights, or training data. By reformulating activation maximization as a semantic optimization problem in text space, \methodname{} combines a trend- and intensity–driven guidance mechanism with an LLM-based refinement process that iteratively constructs natural-language descriptors for a target class. A global and local optimization context further stabilizes the search, enabling the recovery of coherent and persistent semantic cues.

Extensive experiments demonstrate that \methodname{} achieves high semantic class fidelity, reliably identifying the concepts associated with a target class from output probabilities alone. The method also recovers latent training semantics, uncovering features genuinely grounded in the data distribution and reflected in the classifier’s visual decision patterns. Despite its strict black-box setting, \methodname{} matches or surpasses state-of-the-art white-box and grey-box dissection tools across multiple architectures, including CNNs and Vision Transformers.

Beyond semantic recovery, \methodname{} provides a principled mechanism for bias discovery. By extracting fine-grained contextual descriptors and using them to construct slice-aware pseudo-labels, the method supports effective debiasing through standard robust optimization techniques. On Waterbirds and CelebA, the discovered descriptors yield worst-group accuracies competitive with or approaching methods that require full access to model internals or training data, and in some cases rival the performance of oracle group labels.

Taken together, these findings show that meaningful model dissection and auditing are possible even when models are accessible only through an inference API. \methodname{} offers a general semantic lens for inspecting the behaviour, implicit training signals, and systematic biases of modern visual classifiers, opening the door to new forms of transparent, data-free, and architecture-agnostic model analysis.

\section*{Data availability}
All datasets used in this study are publicly available, widely recognised benchmark datasets that are maintained by their respective communities. These standard datasets can be accessed from the following repositories/links: ImageNet~\cite{deng2009imagenet} at \url{https://image-net.org/download.php}, WaterBirds~\cite{sagawadistributionally,WahCUB_200_2011} at \url{https://github.com/kohpangwei/group_DRO}, \url{https://www.vision.caltech.edu/datasets/cub_200_2011/}, CelebA~\cite{liu2015deeplearningfaceattributes} at \url{https://mmlab.ie.cuhk.edu.hk/projects/CelebA.html}. No new data were generated in this study beyond the use of these existing open-source benchmark datasets. All data analysed during this study are therefore openly accessible to readers and researchers.

\section*{Acknowledgments}
The work was partially supported by the Italian Ministerial grants PRIN 2022 “B-Fair: Bias-Free Artificial Intelligence methods for automated visual Recognition”, CUP
E53D23008030006.

\bibliography{sn-bibliography}% common bib file

\end{document}